\pdfoutput=1
\documentclass[journal,twoside,print]{ieeecolor}
\usepackage{generic}
\usepackage{cite}
\usepackage{amsmath,amssymb,amsfonts}
\usepackage{graphicx}
\usepackage{booktabs}
\usepackage{multirow}
\usepackage{tabularx}
\usepackage[table]{xcolor}
\definecolor{clsred}{RGB}{231,76,60}
\definecolor{clsgreen}{RGB}{46,204,113}
\definecolor{clsblue}{RGB}{52,152,219}
\definecolor{clsorange}{RGB}{255,187,120}
\usepackage{caption}
\usepackage{placeins}
\usepackage{algorithm}
\usepackage{algorithmic}
\usepackage{textcomp}
\usepackage{url}
\usepackage{hyperref}
\hypersetup{
  hidelinks,
  pdftitle={Vision--Language Enhanced Foundation Model for Semi-Supervised Medical Image Segmentation},
  pdfauthor={Jiaqi Guo, Mingzhen Li, Hanyu Su, Keigo Healy, Lexiaozi Fan, Neda Tavakoli, Santiago Lopez-Tapia, Daniel Kim, and Aggelos K. Katsaggelos},
  pdfsubject={Preprint submitted to IEEE Transactions on Medical Imaging},
  pdfkeywords={medical image segmentation, semi-supervised learning, vision-language models, foundation models, pseudo-labeling}
}
\usepackage{orcidlink}
\newcommand{\eg}{e.g.}

\providecommand{\refname}{References}
\def\BibTeX{{\rm B\kern-.05em{\sc i\kern-.025em b}\kern-.08em
    T\kern-.1667em\lower.7ex\hbox{E}\kern-.125emX}}

\newcommand{\IEEEpreprintnotice}{This work has been submitted to the IEEE for possible publication. Copyright may be transferred without notice, after which this version may no longer be accessible.}



\makeatletter
\def\ps@titlepagestyle{%
  \def\@oddfoot{}%
  \def\@evenfoot{}%
  \def\@oddhead{%
    \begin{tabular}{@{}p{43pc}@{}}\\[-24pt]
      \vbox{\hsize43pc\scriptsize\textsf{\leftmark}\hfill\textsf{\thepage}\hbox{}}\\[-19pt]
      \vbox{\color{subsectioncolor}\hrule height1pt width43pc depth0pt}%
    \end{tabular}}%
  \def\@evenhead{%
    \begin{tabular}{@{}p{43pc}@{}}\\[-24pt]
      \vbox{\hsize43pc\scriptsize\textsf{\thepage}\hfill\textsf{\leftmark}\hbox{}}\\[-19pt]
      \vbox{\color{subsectioncolor}\hrule height1pt width43pc depth0pt}%
    \end{tabular}}%
}
\makeatother

\begin{document}
\bstctlcite{TMI:BSTcontrol}

\title{Vision--Language Enhanced Foundation Model for Semi-Supervised Medical Image Segmentation}

\author{%
Jiaqi Guo~\orcidlink{0009-0002-6263-2858},
Mingzhen Li~\orcidlink{0009-0007-8381-3216},
Hanyu Su~\orcidlink{0009-0000-4341-1342},
Keigo Healy,
Lexiaozi Fan~\orcidlink{0000-0002-3714-5372},
Neda Tavakoli~\orcidlink{0000-0002-1541-5917},
Santiago López-Tapia~\orcidlink{0000-0003-2090-7446},
Daniel Kim~\orcidlink{0000-0003-2660-8973},
and Aggelos K. Katsaggelos~\orcidlink{0000-0003-4554-0070},
~\IEEEmembership{Fellow,~IEEE}
}

\maketitle

\begin{abstract}
Semi-supervised learning (SSL) has emerged as an efficient paradigm for medical image segmentation, reducing the reliance on extensive expert annotations. Vision–language models (VLMs) have demonstrated strong generalization and few-shot capabilities across diverse visual domains. In this work, we integrate a VLM into a semi-supervised medical image segmentation model by adding a Vision–Language Enhanced Semi-supervised Segmentation Assistant (VESSA) that incorporates foundation-level visual–semantic understanding into SSL frameworks. Our approach consists of two stages. In Stage 1, the VLM-enhanced segmentation foundation model VESSA is trained as a reference-guided segmentation assistant using a template bank containing gold-standard exemplars, simulating learning from limited labeled data. Given an input–template pair, VESSA performs visual feature matching to extract representative semantic and spatial cues from exemplar segmentations, generating structured prompts for a Segment Anything Model (SAM)-inspired mask decoder to produce segmentation masks. In Stage 2, VESSA is integrated into an SSL framework as a plug-and-play teacher, providing template-guided pseudo-labels that complement the task-specific student model and strengthen supervision under scarce annotations. Extensive experiments across multiple segmentation datasets and domains show that VESSA-augmented SSL significantly enhances segmentation accuracy, outperforming state-of-the-art baselines under extremely limited annotation conditions.

\end{abstract}


\par\vfill
\begingroup
\footnotesize\sffamily\sloppy
\noindent\textit{\IEEEpreprintnotice}\par
\noindent Jiaqi Guo and Mingzhen Li are co-first authors and are listed in alphabetical order.\par
\noindent Jiaqi Guo, Mingzhen Li, Hanyu Su, Keigo Healy, Santiago López-Tapia, and Aggelos K. Katsaggelos are with the Department of Electrical and Computer Engineering, Northwestern University, Evanston, IL 60208 USA. (e-mail: jiaqi.guo@northwestern.edu; mingzhenli2029@u.northwestern.edu; hanyusu2026@u.northwestern.edu; keigohealy2028@u.northwestern.edu; santiago.lopeztapia@northwestern.edu; a-katsaggelos@northwestern.edu)\par
\noindent Lexiaozi Fan, Neda Tavakoli, and Daniel Kim are with the Department of Radiology, Northwestern University, Chicago, IL 60611 USA. (e-mail: lexiaozi.fan@northwestern.edu; neda.tavakoli@northwestern.edu; daniel.kim3@northwestern.edu).\par
\endgroup
\newpage

\section{Introduction}
\label{sec:intro}

\begin{figure*}[t]
    \centering
    \includegraphics[width=0.91\textwidth]{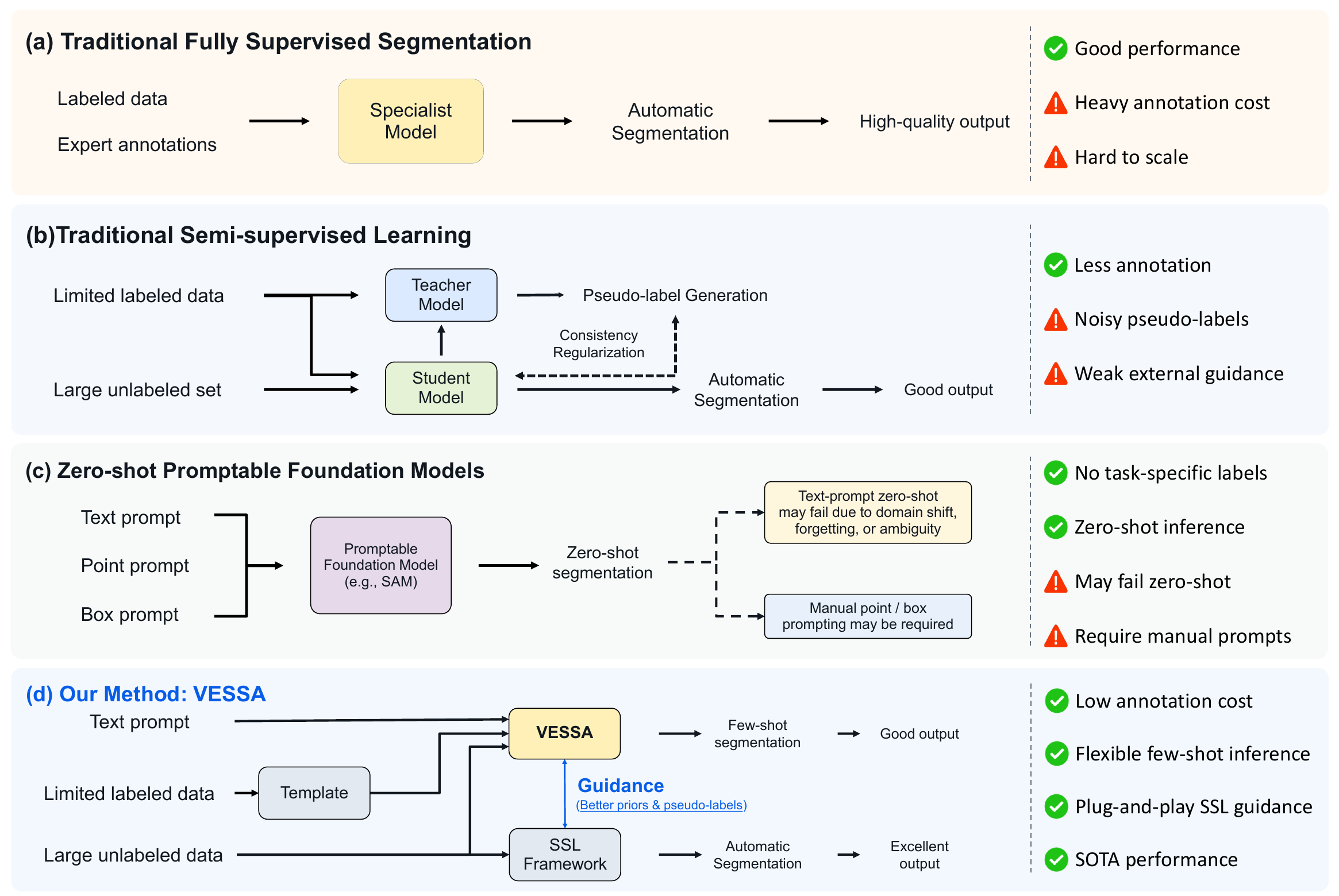}
    \caption{\textbf{Motivation and comparison of segmentation paradigms.} \textbf{(a)} Traditional fully supervised segmentation. \textbf{(b)} Traditional semi-supervised learning. \textbf{(c)} Zero-shot promptable foundation models. \textbf{(d)} Our method: VESSA.}
    \label{fig:motivation}
\end{figure*}

Accurate medical image segmentation supports quantitative diagnosis and treatment planning, but dense expert annotation is costly and difficult to scale across modalities and tasks~\cite{litjens2017survey,wang2022medical,ronneberger2015u,isensee2018nnu,menze2014multimodal,bernard2018deep, lin2024usformer}. Semi-supervised learning (SSL) reduces this burden by combining a small labeled set with unlabeled images through pseudo-labeling and consistency regularization~\cite{yang2023revisiting,yu2019uncertainty,niu2024survey,li2020shape,han2024deep,yang2022survey,tarvainen2017mean}. However, teacher and student models commonly share architectures and training signals, so erroneous pseudo-labels can reinforce confirmation bias, particularly for small or low-contrast structures~\cite{yu2019uncertainty,li2020shape,luo2022semi,arazo2020pseudo,cascante2021curriculum,kwon2022semi}.

Promptable segmentation foundation models, including the Segment Anything Model (SAM) family and its variants, offer a promising route toward label-efficient, generalizable mask generation in few-shot or zero-shot settings from points, boxes, or coarse masks, yet still rely on careful, case-specific prompt engineering~\cite{kirillov2023segment,ravi2024sam,ma2024segment}. More recently, SAM3~\cite{carion2026sam} introduced text prompting as a step toward fully automated segmentation. However, despite reporting that 13.8\% of its training data are medical images, SAM3 remains unreliable for zero-shot medical segmentation, particularly out of domain. This limitation may arise from imbalanced pre-training data, which can weaken performance on underrepresented conditions. Meanwhile, medical SAM adaptations typically require spatial prompts such as points, boxes, or masks. Thus, neither text-only inference nor manual spatial prompting yet provides reliable, fully automated medical segmentation.

These challenges motivate a different setting: instead of relying purely on zero-shot inference, we target realistic few-shot medical segmentation. Although dense segmentation labels are scarce and labor-intensive to collect, many clinical and research workflows have access to a small number of trusted gold-standard masks. Such examples contain valuable anatomical, spatial, and appearance priors that are not captured by text alone. We therefore ask whether a medical segmentation foundation model can combine flexible text prompts with a small bank of gold-standard templates to achieve reliable few-shot inference and to provide stronger supervision for downstream SSL.

To this end, we introduce \textbf{VESSA}, \textbf{V}ision--language \textbf{E}nhanced \textbf{S}emi-supervised \textbf{S}egmentation \textbf{A}ssistant. In Stage~1, VESSA is pre-trained for reference-guided medical image segmentation. Given a target image, a text prompt, and a retrieved gold-standard template, its vision--language backbone and template-embedded memory generate segmentation masks. This design combines semantic flexibility from text with spatial and anatomical guidance from a small template bank. In Stage~2, VESSA serves as a plug-and-play teacher for semi-supervised segmentation. Labeled samples form the template bank, from which VESSA generates pseudo-labels for unlabeled images to complement task-specific SSL predictions. We validate this integration with UniMatch~v1/v2~\cite{yang2023revisiting,yang2025unimatch}, Mean Teacher~\cite{tarvainen2017mean}, and FixMatch~\cite{sohn2020fixmatch}, demonstrating its compatibility with different SSL frameworks. Figure 1 summarizes the compared segmentation paradigms.

In summary, this paper makes the following contributions:
\begin{itemize}
    \item We introduce VESSA, a vision--language enhanced foundation model for few-shot medical image segmentation with flexible text prompts and few gold-standard templates; its reference-based prompting and template-embedded memory emulate realistic test-time scenarios with only a small bank of trusted template images and masks, and training uses 13 public medical datasets spanning multiple modalities and clinical domains.
    \item We propose a systematic pipeline for integrating VESSA into standard semi-supervised medical image segmentation frameworks; by exploiting its template bank and memory mechanisms during inference, VESSA provides strong external supervision from limited labels, generates higher-quality pseudo-labels, improves multiple SSL baselines on public benchmarks, and advances label-efficient medical image segmentation.
\end{itemize}

\begin{figure*}[ht]
    \centering
    \includegraphics[width=0.97\textwidth]{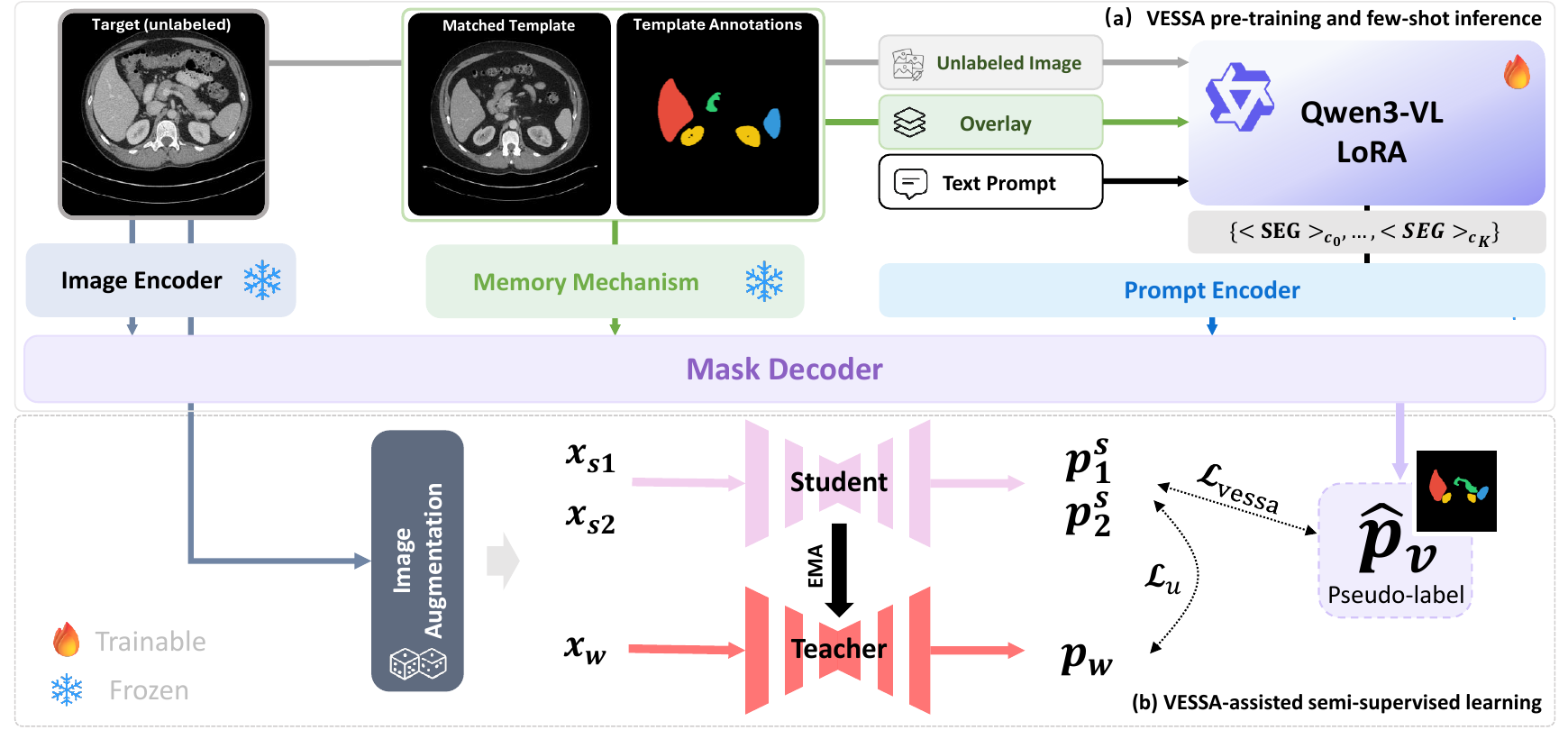}
    \caption{\textbf{Overview of VESSA.}
\textbf{(a) VESSA pre-training and few-shot inference:} The target (unlabeled) image, matched template overlay, and text prompt generate class-aligned segmentation prompts and template-memory features for mask decoding.
\textbf{(b) VESSA-assisted semi-supervised learning:} A student--teacher SSL network enforces weak-to-strong consistency, with VESSA providing template-guided pseudo-labels \(\hat{p}_v\).}
    \label{fig: overall}
\end{figure*}

\section{Related Work}
\label{sec:related-work}

\subsection{Semi-supervised Medical Image Segmentation}
Semi-supervised learning has emerged as a promising paradigm in medical image analysis, addressing the scarcity of expert-annotated data by leveraging both labeled and unlabeled samples for model training~\cite{luo2022semi,han2024deep,yang2023revisiting,li2020shape,wang2022medical,yu2019uncertainty, xie2022learn}. The two commonly adopted strategies for leveraging unlabeled data are pseudo-labeling~\cite{arazo2020pseudo,li2020shape,hung2018adversarial,yu2019uncertainty,sohn2020fixmatch,chen2021semi,huang2023source} and consistency learning~\cite{luo2022semi,laine2016temporal,tarvainen2017mean,verma2022interpolation,miyato2018virtual}. Pseudo-labeling assigns model-generated labels to unlabeled images to expand the training set but is sensitive to early prediction noise. Consistency learning enforces prediction stability under various perturbations to improve robustness but depends heavily on the quality of the supervising signals. Semi-supervised learning for medical image segmentation has generally followed the same principles as in natural-image SSL, relying on pseudo-labeling, consistency regularization, and teacher–student architectures. Early efforts explored adversarial training~\cite{hung2018adversarial,goodfellow2020generative} to refine pseudo-labels or to align the prediction distributions between labeled and unlabeled data. UniMatch~\cite{yang2023revisiting} revisits FixMatch-style~\cite{sohn2020fixmatch} consistency for semantic segmentation by using predictions from weakly augmented views to supervise strongly augmented ones, achieving substantial improvements on both natural and medical datasets. UniMatch~v2~\cite{yang2025unimatch} further upgrades the encoder to ViT/DINOv2~\cite{oquab2023dinov2} and simplifies the augmentation pipeline. Recent medical SSL approaches~\cite{zhang2024semisam,zhang2025semisam+,zhu2025multi} incorporate multi-view or multi-scale consistency constraints, adversarial perturbations, and uncertainty-aware regularization to improve robustness under scarce annotations, enabling models to extract more reliable supervisory signals from unlabeled data.

\subsection{Promptable Segmentation Foundation Models and Their Roles in Semi-Supervised Learning}
Recent SSL studies leverage promptable foundation models such as SAM, SAM2, and their medical adaptations~\cite{kirillov2023segment,ravi2024sam,ma2024segment,zhang2023enhancing}. SemiSAM~\cite{zhang2024semisam} and SemiSAM+~\cite{zhang2025semisam+} use SAM-assisted consistency or pseudo-labeling, KnowSAM~\cite{huang2024knowsam} distills SAM knowledge through learnable prompts, and SFR~\cite{li2024sfr} fine-tunes SAM before training a task-specific 3D segmenter. These methods commonly depend on specialist-generated prompts or setting-specific SAM fine-tuning, making them sensitive to prompt quality. VESSA instead learns a text-prompted, reference-guided interface in which each requested class is associated with text, a segmentation token, template evidence, and memory features. It supports multi-class semantic segmentation and can provide external template-guided pseudo-labels to UniMatch~v2, Mean Teacher, FixMatch, or other SSL frameworks.

\subsection{Vision--Language Guidance for Segmentation and SSL}
Vision--language models (VLMs) have shown that language supervision can provide semantic priors beyond fixed category labels~\cite{radford2021learning,luddecke2022image,lai2024lisa, Li2026_475d05f5, li2025rau}. Previous studies have shown that language cues can help distinguish semantically distinct objects with similar shapes and anatomical features during segmentation. In semantic segmentation, language-guided methods such as LISA~\cite{lai2024lisa} use a special \texttt{<SEG>} token to connect multimodal reasoning with mask decoding, demonstrating that text can serve as a flexible interface for dense prediction. Recent SSL methods also explore vision--language guidance. SemiVL~\cite{hoyer2023semivl} integrates VLM priors into semi-supervised semantic segmentation through spatial fine-tuning and a language-guided decoder. In medical imaging, DuSSS~\cite{pan2025dusss} uses dual semantic similarity supervision to improve cross-modal consistency and refine pseudo-labels, while VCLIPSeg~\cite{li2024vclipseg} incorporates voxel-wise CLIP-enhanced prompts, vision--text consistency, and dynamic pseudo-labeling for semi-supervised medical segmentation. These works demonstrate that language and VLM priors can improve label-efficient segmentation, but they differ from VESSA in both objective and interface. VESSA is not only a VLM-regularized SSL model but also a reference-guided medical segmentation foundation model that combines text prompts with a gold-standard template bank and a template-embedded memory mechanism. This design allows VESSA to operate as a few-shot segmentation model and also as a plug-and-play teacher or pseudo-label generator for different SSL frameworks.

\section{Method}
\label{sec:method}

\noindent\textbf{Overview.} Our framework has two stages, as shown in Fig.~\ref{fig: overall}. In \textbf{Stage 1}, we pre-train \textbf{VESSA} as a vision--language enhanced, reference-guided medical segmentation foundation model. In \textbf{Stage 2}, VESSA is inserted into semi-supervised medical image segmentation frameworks as a plug-and-play teacher or pseudo-label generator.

\subsection{Pre-training of VESSA}
\label{sec:pretraining}

\begin{figure*}[htbp]
    \centering
    \includegraphics[width=0.96\textwidth]{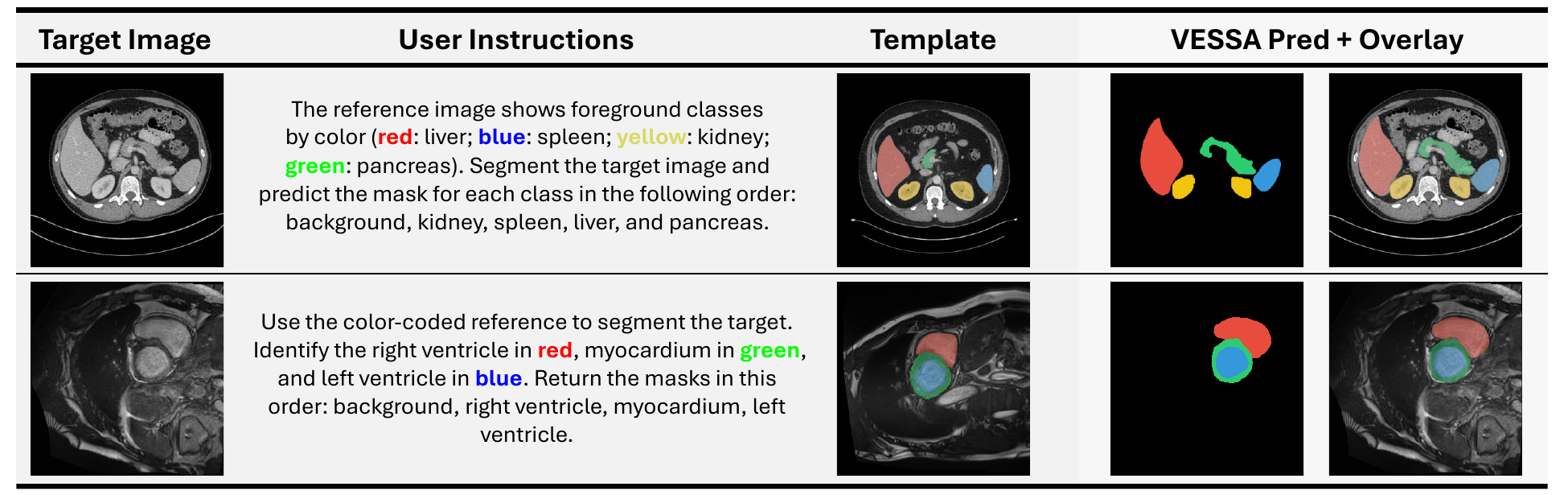}
    \caption{\textbf{Three-component prompts sent to VESSA.} From left to right, the target image, the text prompt, and a template sample used to guide segmentation. Templates are automatically selected through a matching mechanism. The right panels show example segmentation outputs.}
    \label{fig: VESSA EXAPLE}
\end{figure*}

\paragraph{Problem formulation}
Let \(x\in\mathbb{R}^{H\times W\times D}\) denote an input medical image, where \(D\) is the number of image channels. A text prompt specifies a set of requested target classes \(\mathcal{C}=\{c_1,c_2,\ldots,c_K\}\). The goal is to predict a multi-class semantic segmentation map \(\hat{y}\in\{0,1,\ldots,K\}^{H\times W}\), where label 0 denotes background and labels \(1,\ldots,K\) correspond to the requested classes. Equivalently, the model predicts class probabilities \(\hat{P}\in[0,1]^{(K+1)\times H\times W}\), and the final label map is obtained by \(\hat{y}_{u}=\arg\max_{c\in\{0,\ldots,K\}} \hat{P}_{c,u}\) for pixel \(u\).

\paragraph{Template bank and reference-based prompt}
VESSA is trained and validated on public medical segmentation datasets spanning diverse modalities and clinical domains. To simulate few-shot use, we randomly sample 5\%, 10\%, or 20\% of labeled cases per dataset by patient to form a template bank; the effect of template-bank size is analyzed in Section~\ref{sec:sota_comparison} and Table~\ref{tab:vessa_stagewise_performance}. Each template comprises an image \(x_{\mathrm{temp}}^i\) and its gold-standard semantic mask \(y_{\mathrm{temp}}^i\). For retrieval, DINOv2~\cite{oquab2023dinov2} encodes each template image as \(f_i\), yielding \(\mathcal{B}=\{(f_i,x_{\mathrm{temp}}^i,y_{\mathrm{temp}}^i)\}\).

For target image \(x\), we compute its DINOv2 embedding \(f_x\) and select the template with the highest cosine similarity:
\begin{equation}
i^* = \operatorname*{arg\,max}_{i} \cos(f_x,f_i).
\label{eq:template_sampling}
\end{equation}
The selected reference is \((x_{\mathrm{temp}}^{*}, y_{\mathrm{temp}}^{*})=(x_{\mathrm{temp}}^{i^*}, y_{\mathrm{temp}}^{i^*})\); we omit the superscript \(*\) below. This matching strategy pairs each target with the most visually and anatomically relevant template.

For each training sample, VESSA receives the target image, the matched template image with mask overlays, and a reference text prompt (Fig.~\ref{fig: VESSA EXAPLE}). For the requested class set \(\mathcal{C}=\{c_1,\ldots,c_K\}\), template labels share the target semantic space, explicitly aligning each template class \(c_k\) with its target-image counterpart. The template therefore provides class-specific spatial and appearance evidence, while the text prompt supplies semantic guidance for the full target set.

\paragraph{Vision--language encoding and class-aligned \texttt{<SEG>} tokens}
The VLM receives the target image \(x\), the selected template overlay \(x_{\mathrm{overlay}}\), and the input text prompt \(y^{\mathrm{in}}_{\mathrm{text}}\), and generates the text output \(\hat{y}^{\mathrm{txt}} = F_{\mathrm{VLM}}(x,x_{\mathrm{overlay}},y^{\mathrm{in}}_{\mathrm{text}})\).
Following LISA-style segmentation-by-token designs~\cite{lai2024lisa}, we expand the VLM vocabulary with a special token \texttt{<SEG>}. Let \(c_0\) denote background. The model produces \(K{+}1\) positionally aligned tokens \(\{\texttt{<SEG>}_{c_0},\ldots,\texttt{<SEG>}_{c_K}\}\), one for background and one for each requested foreground class in \(\mathcal{C}\). The final-layer embeddings of these tokens are projected by an MLP into segmentation prompt embeddings:
\begin{equation}
 h_k = \mathrm{MLP}\!\left(\mathrm{Emb}_{\mathrm{VLM}}(\texttt{<SEG>}_{c_k})\right), \quad k=0,\ldots,K.
\end{equation}
These embeddings preserve the positional and semantic correspondence between background or each requested foreground class, its text description, and its template evidence.

\paragraph{Template-embedded memory encoding}
In parallel, SAM2-style visual features \(f_{\mathrm{img}} = E_{\mathrm{vision}}(x)\) are extracted from the target image using a vision encoder~\cite{ravi2024sam}. The selected template image and its class-specific or semantic mask are passed to the memory encoder to obtain template memory features \(f_{\mathrm{memory}} = E_{\mathrm{memory}}(x_{\mathrm{temp}}, y_{\mathrm{temp}})\).
The memory features encode visual appearance, spatial priors, and class-specific anatomical context from the template bank. Target image features then attend to the template memory through cross-attention,
\begin{equation}
q = \mathrm{Attention}(f_{\mathrm{img}}, f_{\mathrm{memory}}),
\end{equation}
so that decoding is guided not only by the text prompt but also by gold-standard template evidence.

\paragraph{Multi-class mask decoding}
The projected class-aligned token embeddings \(\{h_k\}_{k=0}^{K}\), image features \(f_{\mathrm{img}}\), and template memory features \(q\) are fused by the prompt encoder and mask decoder. For each class channel \(c_k\), the corresponding prompt embedding is \(p_k = E_{\mathrm{prompt}}(h_k, f_{\mathrm{img}}, q)\), and the decoder predicts a multi-channel logit map:
\begin{equation}
z = D_{\mathrm{mask}}(\{p_k\}_{k=0}^{K}, q), \qquad \hat{P}=\mathrm{softmax}(z).
\label{eq:pred_softmax}
\end{equation}
Here \(z\in\mathbb{R}^{(K+1)\times H\times W}\) contains background and foreground class channels. The final semantic segmentation map is obtained by an argmax over the class dimension.

\paragraph{Training objective}
VESSA is trained with a multi-task objective that supervises both text generation and segmentation:
\begin{equation}
\mathcal{L} = \lambda_{\mathrm{txt}} \mathcal{L}_{\mathrm{txt}} + \lambda_{\mathrm{seg}} \mathcal{L}_{\mathrm{seg}}.
\label{eq:total_loss}
\end{equation}
The language modeling loss is
\begin{equation}
\mathcal{L}_{\mathrm{txt}} = -\sum_{t=1}^{T}\log P(\hat{y}^{\mathrm{txt}}_t = y^{\mathrm{txt}}_t),
\end{equation}
where \(y^{\mathrm{txt}}_t\) is the target token at timestep \(t\). The segmentation branch is supervised as a single \emph{multi-class} problem over the \(K{+}1\) softmax channels \(\hat{P}=\mathrm{softmax}(z)\) of Eq.~\eqref{eq:pred_softmax} (\(K\) foreground classes plus background), combining a class-weighted multi-class cross-entropy and a multi-class Dice term:
\begin{equation}
\mathcal{L}_{\mathrm{seg}} = \lambda_{\mathrm{ce}}\,\mathcal{L}^{\mathrm{seg}}_{\mathrm{CE}} + \lambda_{\mathrm{dice}}\,\mathcal{L}^{\mathrm{multi}}_{\mathrm{Dice}}.
\label{eq:mask_loss}
\end{equation}
Let \(Y\in\{0,1\}^{(K+1)\times H\times W}\) be the one-hot ground truth and \(y_u\in\{0,\ldots,K\}\) the ground-truth class index at pixel \(u\). The class-weighted multi-class cross-entropy over the softmax probabilities is
\begin{equation}
\mathcal{L}^{\mathrm{seg}}_{\mathrm{CE}}
= -\frac{1}{\sum_{u\in\Omega} w_{y_u}}\sum_{u\in\Omega} w_{y_u}\,\log \hat{P}_{y_u,u},
\end{equation}
where \(w_c\) weights class \(c\); we set \(w_0=0.2\) for background and \(w_c=1\) for each foreground class to counter background dominance. The multi-class Dice loss is
\begin{equation}
\mathcal{L}^{\mathrm{multi}}_{\mathrm{Dice}} = 1 - \frac{1}{K+1}\sum_{k=0}^{K}
\frac{2\sum_{u} Y_{k,u}\hat{P}_{k,u}+\epsilon}
{\sum_{u}Y_{k,u}+\sum_{u}\hat{P}_{k,u}+\epsilon},
\end{equation}
where \(\Omega\) is the pixel set and \(\epsilon\) is a small constant for numerical stability. At inference the label map is obtained by \(\hat{y}_u=\arg\max_{c}\hat{P}_{c,u}\); Dice and mIoU are then reported over the foreground classes only.

\subsection{VESSA in Semi-Supervised Medical Image Segmentation}
\label{sec:framework}

\paragraph{Preliminaries}
We adopt UniMatch~v2~\cite{yang2025unimatch} as the task-specific SSL framework. Given labeled and unlabeled sets \(\mathcal{D}_l=\{(x_i^l,y_i^l)\}\) and \(\mathcal{D}_u=\{x_i^u\}\), the student \(f_{\theta_s}\) is trained on labeled data with \(\mathcal{L}_{\mathrm{sup}}=\mathcal{L}^{\mathrm{multi}}_{\mathrm{CE}}(f_{\theta_s}(x_i^l),y_i^l)\). For each unlabeled image, an EMA teacher predicts a weakly augmented view; high-confidence pixels whose confidence is at least \(\tau\) yield hard pseudo-labels that supervise the student on two strongly augmented views through the original weak-to-strong consistency loss \(\mathcal{L}_{\mathrm{u}}\).

\paragraph{VESSA-assisted pseudo-labeling}
At the beginning of downstream SSL training, the labeled images and their masks are inserted into the VESSA template bank. For each unlabeled image \(x^u\), VESSA retrieves the most similar template according to Eq.~\ref{eq:template_sampling}. The target image, selected template overlay, and text prompt specifying the requested classes \(\mathcal{C}\) are then passed to VESSA to produce multi-class probabilities
\begin{equation}
p_v = f_{\mathrm{VESSA}}(x^u,x_{\mathrm{temp}},y^{\mathrm{in}}_{\mathrm{text}}),
\end{equation}
and a VESSA pseudo-label \(\hat{p}_v=\arg\max_c p_{v,c}\). Because VESSA uses text and gold-standard template evidence, \(\hat{p}_v\) provides an external supervision signal complementary to the EMA teacher.

\paragraph{VESSA distillation loss}
Let \(p^{s}\) denote the student prediction on a strongly augmented view of an unlabeled image. VESSA supervises the student through a distillation loss that combines a hard cross-entropy term against the VESSA pseudo-label \(\hat{p}_v\) and a soft Kullback--Leibler (KL) term against the full VESSA class-probability map \(p_v\):
\begin{equation}
\mathcal{L}_{\mathrm{vessa}} = (1-\beta)\,H(p^{s},\hat{p}_v) + \beta\,\mathrm{KL}\!\left(p_v \,\big\|\, p^{s}\right),
\label{eq:lvessa}
\end{equation}
where \(H(\cdot,\cdot)\) is multi-class cross-entropy and \(\beta\in[0,1]\) balances the hard-label and soft-distillation terms. The soft KL term preserves the full class-probability information in VESSA's predictions rather than only its top-1 label.

\paragraph{Dynamic weight scheduling}
The EMA teacher is often unreliable early in training but improves as the student learns task-specific decision boundaries, whereas VESSA provides stable template-guided predictions from the beginning. We therefore decay the influence of the VESSA guidance over training: the VESSA distillation term is weighted by \(w_v(t)\), which follows a cosine schedule from a positive initial value to zero, while the supervised and consistency weights are held fixed. Early training thus emphasizes template-guided VESSA supervision, and later training increasingly relies on the task-adapted student and its EMA teacher.

\paragraph{Overall training loss}
The final Stage-2 objective combines the supervised loss on labeled data, the UniMatch~v2 consistency loss \(\mathcal{L}_{\mathrm{u}}\) on unlabeled data, and the VESSA distillation loss:
\begin{equation}
\mathcal{L}= w_x\,\mathcal{L}_{\mathrm{sup}} + w_u\,\mathcal{L}_{\mathrm{u}} + w_v(t)\,\mathcal{L}_{\mathrm{vessa}},
\label{eq:total_joint_loss}
\end{equation}
where \(\mathcal{L}_{\mathrm{u}}\) is the weak-to-strong consistency loss and \(w_x, w_u\) are fixed weights. The same integration principle can be applied to other semi-supervised frameworks, including Mean Teacher, FixMatch and UniMatch~v1, by augmenting their self-generated pseudo-labels with VESSA's multi-class template-guided pseudo-labels.

\section{Experiments}
\label{sec:experiments}

This section evaluates VESSA as a transferable reference-guided foundation model in Stage~1 and a plug-and-play pseudo-labeling teacher for scarce-label semi-supervised learning (SSL) in Stage~2.

\subsection{Experimental Setup}
\textbf{Stage-1 VESSA training and validation datasets.}
VESSA is first trained as a reference-guided medical segmentation foundation model on a diverse collection of public segmentation datasets spanning CT, MRI, ultrasound, and chest X-ray modalities. The Stage-1 training/validation pool covers diverse anatomical regions and clinical tasks, including abdominal organs, cardiac structures, brain tumors, breast ultrasound lesions, lung fields, and dental structures. This stage trains VESSA to integrate text prompts, matched templates, and template-embedded memory for reference-guided semantic segmentation. To evaluate whether the learned foundation model transfers beyond its Stage-1 distribution, we separate the downstream Stage-2 SSL datasets from the Stage-1 training data.

\textbf{Stage-2 SSL and evaluation datasets.}
The downstream evaluation spans five datasets that deliberately cover heterogeneous modalities, anatomies, and task difficulties: \textbf{ACDC}, \textbf{AbdomenCT-1K}, \textbf{BUSI}, \textbf{STS-2D Tooth}, and \textbf{Shenzhen CXR}. \textbf{ACDC}~\cite{bernard2018deep} comprises short-axis cine cardiac MRI from 100 annotated patients spanning a healthy group and four pathological groups, with three foreground classes: the left-ventricular cavity, right-ventricular cavity, and left-ventricular myocardium, delineated at the end-diastolic and end-systolic phases. \textbf{AbdomenCT-1K}~\cite{ma2021abdomenct} is a large multi-center abdominal CT benchmark of over 1{,}000 scans in which the liver, kidney, spleen, and pancreas are uniformly annotated. \textbf{BUSI} is the Breast Ultrasound Images dataset~\cite{aldhabyani2020busi} of 780 B-mode scans collected from 600 women (benign, malignant, and normal cases); following our multi-class protocol we jointly segment the lesion and distinguish its benign versus malignant category. \textbf{STS-2D Tooth}~\cite{sts2d2024} provides dental panoramic X-ray images with expert-delineated tooth masks, drawn from the 2D task of the Semi-supervised Teeth Segmentation challenge. \textbf{Shenzhen CXR}~\cite{jaeger2014two,stirenko2018chest} consists of 566 frontal chest radiographs (279 normal and 287 with tuberculosis manifestations) collected at Shenzhen No.~3 People's Hospital, China, each annotated with left- and right-lung fields. Together these datasets probe both in-domain performance and transfer to structures, modalities, and domains held out from Stage-1 pre-training; for each, we evaluate semi-supervised learning under 1\%, 5\%, and 10\% labeled-data regimes. Across all experiments, we report Dice and mIoU, which measure region overlap and class-normalized overlap, respectively.

\textbf{Implementation details.}
All models are implemented in PyTorch~2.8 (Python 3.12, CUDA 12.6, cuDNN 9.10.2) and trained on two NVIDIA A100 GPUs (40 or 80\,GB each). VESSA uses Qwen3-VL-4B-Instruct~\cite{bai2025qwen3} as the vision--language model and SAM2.1 (Hiera-Large)~\cite{ravi2024sam} as the segmentation backbone. Template images are encoded offline and retrieved with FAISS using a frozen DINOv2 ViT-S/14~\cite{oquab2023dinov2}. Only Qwen3-VL is adapted with LoRA~\cite{hu2022lora} (rank 64, $\alpha=128$, dropout 0.05) on all attention and MLP projections ($q$, $k$, $v$, $o$, gate, up, and down). Stage~1 uses AdamW ($\beta=(0.9,0.95)$, zero weight decay, gradient clipping at 1.0) with a learning rate of $1\times10^{-5}$ and a linear warmup--decay schedule with 100 warmup steps. Training runs for 6{,}000 steps ($60\times100$) with an effective batch size of 16 (micro-batch size 2 and eight-step gradient accumulation) in bfloat16 using DeepSpeed ZeRO-2~\cite{rasley2020deepspeed}. For Stage~2 SSL, UniMatch~v2~\cite{yang2025unimatch} is the primary student framework, using a DINOv2 encoder and DPT~\cite{ranftl2021vision} decoder under Distributed Data Parallel. We additionally integrate VESSA with Mean Teacher~\cite{tarvainen2017mean} and FixMatch~\cite{sohn2020fixmatch}, and follow official or recommended settings for competing methods when available.

\textbf{Statistical analysis.}
For Table~\ref{tab:comparison}, Dice and mIoU are computed for each held-out patient after aggregating all images, slices, or frames belonging to that patient. Results are reported as the mean $\pm$ standard deviation across patients; thus, the standard deviation reflects inter-patient variability rather than variation across slices, training seeds, labeled-data splits, or repeated runs. The same held-out patient cohort is used for all methods, and the number of observations in each analysis equals the number of patients in the corresponding test cohort. Dice and mIoU are analyzed separately. For each dataset and labeled-data ratio, method is treated as a six-level within-patient factor (five baselines and VESSA) in a repeated-measures ANOVA. Greenhouse--Geisser-adjusted degrees of freedom are used because sphericity is not assumed. Following a significant omnibus test, VESSA is compared with each baseline using paired patient-level $t$-tests; Bonferroni correction is applied to the family of five comparisons separately for each dataset, labeled-data ratio, and metric.

\subsection{Experiment Results}
\label{sec:sota_comparison}
\renewcommand{\arraystretch}{1.05}
\newcommand{\experimenttablefont}{\fontsize{8.5pt}{9.5pt}\selectfont}

\begin{table*}[!t]
\centering
\experimenttablefont
\setlength{\tabcolsep}{4pt}
\caption{Few-shot inference performance of the pre-trained VESSA foundation model on Stage-1 validation datasets and held-out downstream datasets. Stage-1 validation rows evaluate VESSA on datasets seen during Stage-1 pre-training; held-out downstream rows evaluate it on the five downstream datasets held out from pre-training, before any task-specific SSL training. ``No-ref'' is the reference-free (zero-shot) variant; 5/10/20\% denote the reference-template bank size. The Train/Val column reports training and validation sample counts. Each entry reports Dice and mIoU; within each row and metric, the best score is shown in bold and the second-best score is underlined.}
\label{tab:vessa_stagewise_performance}
\begin{tabularx}{\textwidth}{l|>{\centering\arraybackslash}p{0.085\textwidth}|>{\centering\arraybackslash}p{0.09\textwidth}|>{\centering\arraybackslash}p{0.095\textwidth}|>{\centering\arraybackslash}X>{\centering\arraybackslash}X|>{\centering\arraybackslash}X>{\centering\arraybackslash}X|>{\centering\arraybackslash}X>{\centering\arraybackslash}X|>{\centering\arraybackslash}X>{\centering\arraybackslash}X}
\toprule
\multirow{2}{*}{Dataset} & \multirow{2}{*}{Train/Val} & \multirow{2}{*}{Usage} & \multirow{2}{*}{Modality} & \multicolumn{2}{c|}{No-ref} & \multicolumn{2}{c|}{5\%} & \multicolumn{2}{c|}{10\%} & \multicolumn{2}{c}{20\%} \\
\cline{5-12}
 & & & & Dice & mIoU & Dice & mIoU & Dice & mIoU & Dice & mIoU \\
\midrule
Montgomery CXR~\cite{jaeger2014two,candemir2014lung} & 110/28 & \multirow{13}{*}{\centering Validation} & X-ray (CXR) & 0.958 & 0.920 & \underline{0.960} & \underline{0.924} & \textbf{0.964} & \textbf{0.932} & 0.959 & 0.922 \\
BUS-UCLM~\cite{vallez2025bus} & 208/52 &  & Ultrasound & 0.429 & 0.373 & \textbf{0.782} & \textbf{0.680} & 0.723 & 0.630 & \underline{0.780} & \underline{0.677} \\
MSD-Prostate~\cite{antonelli2022medical} & 728/186 &  & MRI & 0.548 & 0.442 & 0.666 & 0.538 & \textbf{0.682} & \textbf{0.560} & \underline{0.667} & \underline{0.543} \\
MSD-Heart~\cite{antonelli2022medical} & 959/248 &  & MRI & 0.839 & 0.748 & 0.856 & 0.768 & \textbf{0.860} & \textbf{0.774} & \underline{0.858} & \underline{0.770} \\
LGG~\cite{buda2019association} & 1014/257 &  & MRI & 0.250 & 0.191 & 0.767 & 0.668 & \textbf{0.788} & \textbf{0.690} & \underline{0.781} & \underline{0.683} \\
M\&Ms~\cite{campello2021multi} & 2880/630 &  & MRI & 0.755 & 0.664 & \underline{0.819} & \underline{0.722} & 0.818 & \textbf{0.726} & \textbf{0.822} & \textbf{0.726} \\
TotalSegmentator-MRI~\cite{dantonoli2025totalsegmentatormri} & 6083/582 &  & MRI & 0.373 & 0.294 & \underline{0.517} & \underline{0.426} & 0.441 & 0.354 & \textbf{0.531} & \textbf{0.437} \\
STS-3D-Tooth~\cite{sts2d2024} & 6840/1747 &  & CT & 0.641 & 0.537 & \underline{0.675} & \underline{0.576} & \textbf{0.678} & \textbf{0.583} & 0.663 & 0.571 \\
AMOS22-MRI~\cite{ji2022amos} & 9128/1579 &  & MRI & 0.401 & 0.342 & \underline{0.665} & \underline{0.578} & 0.652 & 0.572 & \textbf{0.713} & \textbf{0.619} \\
BraTS~\cite{menze2014multimodal} & 13088/3113 &  & MRI & 0.093 & 0.066 & \textbf{0.598} & \underline{0.470} & \underline{0.597} & \textbf{0.471} & 0.588 & 0.463 \\
CAMUS~\cite{leclerc2019deep} & 17068/4164 &  & Ultrasound & 0.858 & 0.760 & \underline{0.870} & \underline{0.776} & \textbf{0.876} & \textbf{0.785} & 0.869 & 0.775 \\
TotalSegmentator-CT~\cite{wasserthal2023totalsegmentator} & 21513/3073 &  & CT & 0.514 & 0.414 & \underline{0.634} & \underline{0.530} & 0.618 & 0.512 & \textbf{0.647} & \textbf{0.544} \\
AMOS22-CT~\cite{ji2022amos} & 30395/8349 &  & CT & 0.310 & 0.272 & \underline{0.651} & \underline{0.586} & \textbf{0.664} & \textbf{0.601} & 0.643 & 0.578 \\
\midrule
Shenzhen CXR~\cite{jaeger2014two,stirenko2018chest} & --/22 & \multirow{5}{*}{\centering \shortstack{Zero-shot/\\Few-shot\\inference}} & X-ray (CXR) & \underline{0.944} & 0.894 & 0.940 & 0.890 & \textbf{0.950} & \textbf{0.907} & \underline{0.944} & \underline{0.896} \\
BUSI~\cite{aldhabyani2020busi} & --/128 &  & Ultrasound & 0.337 & 0.289 & \underline{0.720} & \underline{0.627} & 0.704 & 0.609 & \textbf{0.740} & \textbf{0.637} \\
STS-2D Tooth~\cite{sts2d2024} & --/180 &  & X-ray (dental) & 0.689 & 0.533 & \textbf{0.816} & \textbf{0.692} & 0.788 & 0.653 & \underline{0.812} & \underline{0.686} \\
ACDC~\cite{bernard2018deep} & --/1076 &  & MRI & 0.733 & 0.644 & \underline{0.805} & \underline{0.715} & \textbf{0.809} & \textbf{0.723} & 0.799 & 0.711 \\
AbdomenCT~\cite{ma2021abdomenct} & --/1166 &  & CT & 0.477 & 0.421 & 0.636 & 0.560 & \underline{0.776} & \underline{0.704} & \textbf{0.805} & \textbf{0.732} \\
\bottomrule
\end{tabularx}
\end{table*}

\begin{table*}[!t]
\centering
\experimenttablefont
\setlength{\tabcolsep}{2pt}
\renewcommand{\arraystretch}{1.0}
\newcommand{\stat}[2]{{\fontsize{7pt}{8pt}\selectfont #1{\fontsize{5pt}{6pt}\selectfont$\mkern0.5mu\pm\mkern0.5mu$}#2}}
\caption{Multi-class semantic segmentation performance on five Stage-2 datasets under 1\%, 5\%, and 10\% labeled-data regimes. Values are patient-level mean $\pm$ standard deviation. Within each labeled-data block and metric column, the best score is bold and the second-best is underlined. $^{\dagger}$Significantly higher than all five baselines on both metrics under the patient-level procedure described in the statistical analysis ($p<0.05$).}
\label{tab:comparison}
\begin{tabularx}{\textwidth}{l|>{\centering\arraybackslash}p{0.07\textwidth}|>{\centering\arraybackslash}X>{\centering\arraybackslash}X|>{\centering\arraybackslash}X>{\centering\arraybackslash}X|>{\centering\arraybackslash}X>{\centering\arraybackslash}X|>{\centering\arraybackslash}X>{\centering\arraybackslash}X|>{\centering\arraybackslash}X>{\centering\arraybackslash}X}
\toprule
\multirow{2}{*}{Method} & \multirow{2}{*}{Labeled} & \multicolumn{2}{c|}{ACDC} & \multicolumn{2}{c|}{AbdomenCT} & \multicolumn{2}{c|}{BUSI} & \multicolumn{2}{c|}{STS-2D Tooth} & \multicolumn{2}{c}{Shenzhen CXR} \\
\cline{3-12}
 &  & Dice & mIoU & Dice & mIoU & Dice & mIoU & Dice & mIoU & Dice & mIoU \\
\midrule
BCP & \multirow{6}{*}{\centering 1\%} & \stat{0.756}{0.067} & \stat{0.685}{0.066} & \stat{0.550}{0.150} & \stat{0.490}{0.155} & \underline{\stat{0.299}{0.265}} & \underline{\stat{0.231}{0.231}} & \stat{0.800}{0.144} & \stat{0.685}{0.151} & \textbf{\stat{0.956}{0.015}} & \textbf{\stat{0.916}{0.026}} \\
SemiVL &  & \stat{0.680}{0.065} & \stat{0.584}{0.066} & \underline{\stat{0.715}{0.104}} & \underline{\stat{0.654}{0.102}} & \stat{0.119}{0.207} & \stat{0.088}{0.168} & \stat{0.883}{0.031} & \stat{0.791}{0.047} & \stat{0.820}{0.334} & \stat{0.780}{0.318} \\
KnowSAM &  & \stat{0.704}{0.062} & \stat{0.626}{0.059} & \stat{0.614}{0.112} & \stat{0.552}{0.111} & \stat{0.220}{0.232} & \stat{0.167}{0.190} & \stat{0.868}{0.031} & \stat{0.768}{0.045} & \underline{\stat{0.955}{0.014}} & \underline{\stat{0.914}{0.025}} \\
SemiSAM+ &  & \stat{0.636}{0.076} & \stat{0.541}{0.078} & \stat{0.518}{0.098} & \stat{0.456}{0.097} & \stat{0.150}{0.181} & \stat{0.108}{0.154} & \stat{0.867}{0.058} & \stat{0.769}{0.079} & \stat{0.869}{0.183} & \stat{0.806}{0.208} \\
UniMatch~v2 &  & \underline{\stat{0.786}{0.056}} & \underline{\stat{0.703}{0.060}} & \stat{0.684}{0.093} & \stat{0.632}{0.091} & \stat{0.216}{0.340} & \stat{0.175}{0.288} & \textbf{\stat{0.892}{0.029}} & \textbf{\stat{0.807}{0.046}} & \stat{0.950}{0.034} & \stat{0.907}{0.057} \\
Ours &  & \textbf{\stat{0.822}{0.048}}$^{\dagger}$ & \textbf{\stat{0.736}{0.049}}$^{\dagger}$ & \textbf{\stat{0.745}{0.089}}$^{\dagger}$ & \textbf{\stat{0.685}{0.090}}$^{\dagger}$ & \textbf{\stat{0.525}{0.354}}$^{\dagger}$ & \textbf{\stat{0.447}{0.330}}$^{\dagger}$ & \underline{\stat{0.888}{0.027}} & \underline{\stat{0.800}{0.043}} & \textbf{\stat{0.956}{0.019}} & \textbf{\stat{0.916}{0.032}} \\
\midrule
BCP & \multirow{6}{*}{\centering 5\%} & \stat{0.790}{0.062} & \stat{0.724}{0.061} & \stat{0.705}{0.126} & \stat{0.649}{0.129} & \stat{0.458}{0.359} & \stat{0.389}{0.332} & \stat{0.873}{0.090} & \stat{0.782}{0.101} & \textbf{\stat{0.964}{0.011}} & \textbf{\stat{0.931}{0.020}} \\
SemiVL &  & \stat{0.720}{0.058} & \stat{0.632}{0.056} & \stat{0.764}{0.086} & \stat{0.705}{0.085} & \stat{0.461}{0.389} & \stat{0.389}{0.353} & \underline{\stat{0.907}{0.034}} & \underline{\stat{0.831}{0.054}} & \underline{\stat{0.963}{0.012}} & \underline{\stat{0.928}{0.021}} \\
KnowSAM &  & \stat{0.782}{0.058} & \stat{0.713}{0.055} & \stat{0.766}{0.120} & \stat{0.712}{0.124} & \stat{0.572}{0.367} & \stat{0.495}{0.342} & \stat{0.895}{0.025} & \stat{0.812}{0.039} & \textbf{\stat{0.964}{0.009}} & \textbf{\stat{0.931}{0.016}} \\
SemiSAM+ &  & \stat{0.745}{0.065} & \stat{0.669}{0.064} & \stat{0.701}{0.100} & \stat{0.644}{0.101} & \stat{0.366}{0.304} & \stat{0.301}{0.272} & \stat{0.891}{0.038} & \stat{0.805}{0.058} & \stat{0.960}{0.014} & \stat{0.923}{0.024} \\
UniMatch~v2 &  & \underline{\stat{0.843}{0.039}} & \underline{\stat{0.765}{0.040}} & \underline{\stat{0.831}{0.068}} & \underline{\stat{0.776}{0.071}} & \underline{\stat{0.626}{0.336}} & \underline{\stat{0.542}{0.317}} & \underline{\stat{0.907}{0.022}} & \stat{0.830}{0.036} & \stat{0.958}{0.022} & \stat{0.919}{0.039} \\
Ours &  & \textbf{\stat{0.861}{0.035}}$^{\dagger}$ & \textbf{\stat{0.783}{0.036}}$^{\dagger}$ & \textbf{\stat{0.843}{0.070}}$^{\dagger}$ & \textbf{\stat{0.789}{0.073}}$^{\dagger}$ & \textbf{\stat{0.639}{0.327}} & \textbf{\stat{0.546}{0.306}} & \textbf{\stat{0.912}{0.021}}$^{\dagger}$ & \textbf{\stat{0.838}{0.034}}$^{\dagger}$ & \stat{0.961}{0.013} & \stat{0.925}{0.022} \\
\midrule
BCP & \multirow{6}{*}{\centering 10\%} & \stat{0.800}{0.059} & \stat{0.736}{0.059} & \stat{0.759}{0.122} & \stat{0.706}{0.126} & \stat{0.474}{0.328} & \stat{0.413}{0.307} & \stat{0.898}{0.085} & \stat{0.821}{0.092} & \textbf{\stat{0.966}{0.012}} & \textbf{\stat{0.934}{0.022}} \\
SemiVL &  & \stat{0.733}{0.062} & \stat{0.648}{0.059} & \stat{0.783}{0.084} & \stat{0.727}{0.084} & \stat{0.429}{0.367} & \stat{0.358}{0.332} & \stat{0.906}{0.033} & \stat{0.829}{0.052} & \stat{0.949}{0.043} & \stat{0.906}{0.068} \\
KnowSAM &  & \stat{0.801}{0.054} & \stat{0.735}{0.052} & \stat{0.815}{0.097} & \stat{0.763}{0.101} & \stat{0.616}{0.355} & \stat{0.538}{0.336} & \stat{0.894}{0.027} & \stat{0.809}{0.043} & \stat{0.959}{0.032} & \stat{0.924}{0.052} \\
SemiSAM+ &  & \stat{0.765}{0.061} & \stat{0.692}{0.060} & \stat{0.739}{0.101} & \stat{0.683}{0.101} & \stat{0.507}{0.334} & \stat{0.437}{0.316} & \stat{0.897}{0.029} & \stat{0.815}{0.046} & \stat{0.959}{0.023} & \stat{0.923}{0.040} \\
UniMatch~v2 &  & \underline{\stat{0.857}{0.031}} & \underline{\stat{0.780}{0.031}} & \underline{\stat{0.856}{0.062}} & \underline{\stat{0.802}{0.064}} & \underline{\stat{0.673}{0.336}} & \underline{\stat{0.589}{0.322}} & \textbf{\stat{0.910}{0.025}} & \textbf{\stat{0.835}{0.040}} & \stat{0.961}{0.020} & \stat{0.926}{0.034} \\
Ours &  & \textbf{\stat{0.869}{0.028}}$^{\dagger}$ & \textbf{\stat{0.792}{0.030}}$^{\dagger}$ & \textbf{\stat{0.857}{0.058}} & \textbf{\stat{0.804}{0.062}} & \textbf{\stat{0.700}{0.310}} & \textbf{\stat{0.613}{0.302}} & \underline{\stat{0.907}{0.021}} & \underline{\stat{0.830}{0.034}} & \underline{\stat{0.965}{0.009}} & \underline{\stat{0.932}{0.016}} \\
\bottomrule
\end{tabularx}
\end{table*}

\textbf{Quantitative results.}
Table~\ref{tab:vessa_stagewise_performance} reports few-shot inference performance of the pre-trained VESSA foundation model on in-domain and held-out datasets. Given only a small bank of reference templates, VESSA produces strong text-prompted segmentation across diverse modalities and anatomies, including STS-2D Tooth, a 2D dental panoramic X-ray task unseen during Stage-1 training. Performance improves as the reference-template bank grows from 5\% to 10\% and then largely plateaus, with 20\% offering only marginal and inconsistent gains; we therefore use the 10\% checkpoint in subsequent Stage-2 experiments.

Table~\ref{tab:comparison} summarizes the Stage-2 comparison against representative semi-supervised and foundation-model-assisted baselines. VESSA-assisted SSL (``Ours'') attains the best or near-best Dice and mIoU across the five datasets, with the largest margins in the 1\% labeled setting, where conventional teacher--student pseudo-labeling is most vulnerable to noisy early predictions. At 1\% labeled, it improves Dice over UniMatch~v2 by 3.6 points on ACDC (0.822 vs.\ 0.786), 6.1 points on AbdomenCT (0.745 vs.\ 0.684), and 30.9 points on the difficult BUSI lesion task (0.525 vs.\ 0.216). As the labeled ratio increases to 10\%, the task-specific student becomes more reliable and the gap narrows. By contrast, STS-2D Tooth and Shenzhen CXR have limited headroom because they involve comparatively simple targets; VESSA's largest gains therefore concentrate on ACDC, AbdomenCT, and BUSI, which contain smaller structures, weaker boundaries, or greater inter-subject variability.

A repeated-measures ANOVA with Bonferroni-corrected paired post-hoc tests confirms that these gains are statistically significant where the task is hardest. Using the rounded means displayed in Table~\ref{tab:comparison}, under 1\% supervision VESSA improves over the strongest displayed baseline by $+3.6$~Dice~/~$+3.3$ mIoU on ACDC (\(p<0.001\)), $+3.0$~/~$+3.1$ on AbdomenCT (\(p<0.05\)), and $+22.6$~/~$+21.6$ on BUSI (\(p<0.001\)); the ACDC advantage remains significant at 5\% and 10\% labels. On the near-saturated Shenzhen CXR benchmark, differences are not significant; on STS-2D Tooth, the differences are generally small, with significance observed only under the 5\% labeled setting. Overall, VESSA is significantly best in 7 of the 15 settings on both metrics and is never significantly outperformed.

\begin{figure*}[t]
    \centering
    \includegraphics[width=0.97\textwidth]{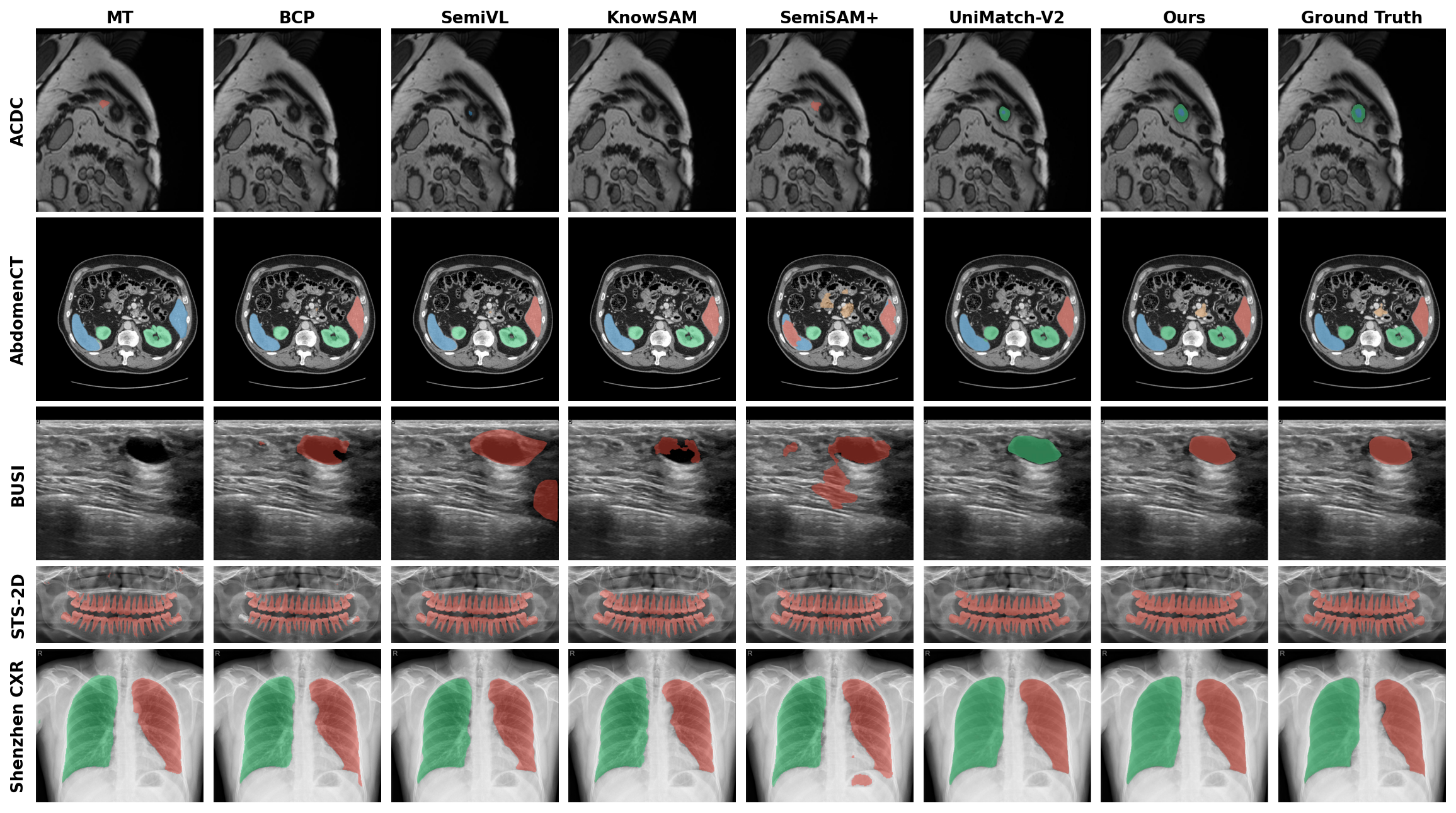}
    \caption{Qualitative comparison on the five Stage-2 datasets under the 1\% labeled setting. Each row shows one representative case (top to bottom: ACDC, AbdomenCT, BUSI, STS-2D Tooth, Shenzhen CXR); the columns compare the semi-supervised and foundation-model-assisted baselines, our VESSA-assisted method, and the ground truth. Foreground classes are color-coded per dataset: ACDC---\textcolor{clsred}{right ventricle}, \textcolor{clsgreen}{myocardium}, \textcolor{clsblue}{left ventricle}; AbdomenCT---\textcolor{clsred}{liver}, \textcolor{clsgreen}{kidney}, \textcolor{clsblue}{spleen}, \textcolor{clsorange}{pancreas}; BUSI---\textcolor{clsred}{benign tumor}, \textcolor{clsgreen}{malignant tumor}; STS-2D Tooth---\textcolor{clsred}{teeth}; Shenzhen CXR---\textcolor{clsgreen}{right lung}, \textcolor{clsred}{left lung}. VESSA-assisted SSL yields more coherent anatomical structures and more precise boundaries than the competing baselines.}
    \label{fig:qual}
\end{figure*}

\begin{table*}[t]
\centering
\fontsize{7.0pt}{7.8pt}\selectfont
\setlength{\tabcolsep}{2pt}
\caption{Performance of VESSA integrated with different SSL frameworks on the five Stage-2 datasets. ``VESSA distill'' directly distills VESSA pseudo-labels without an additional SSL framework. Parentheses show changes from the corresponding framework without VESSA, computed from unrounded scores (rounding may cause differences of 0.001).}
\label{tab:vessa_frameworks}
\resizebox{\textwidth}{!}{%
\begin{tabular}{@{}l|c|cc|cc|cc|cc|cc@{}}
\toprule
\multirow{2}{*}{\shortstack{Framework\\(+ VESSA)}} & \multirow{2}{*}{Labeled} & \multicolumn{2}{c|}{ACDC} & \multicolumn{2}{c|}{AbdomenCT} & \multicolumn{2}{c|}{BUSI} & \multicolumn{2}{c|}{STS-2D Tooth} & \multicolumn{2}{c}{Shenzhen CXR} \\
\cline{3-12}
 &  & Dice & mIoU & Dice & mIoU & Dice & mIoU & Dice & mIoU & Dice & mIoU \\
\midrule
VESSA distill & \multirow{5}{*}{\centering 1\%} & 0.804 & 0.713 & 0.684 & 0.615 & 0.507 & 0.438 & 0.846 & 0.734 & 0.940 & 0.889 \\
MT &  & \mbox{0.804\,(\textcolor{red}{$\uparrow$}0.015)} & \mbox{0.717\,(\textcolor{red}{$\uparrow$}0.022)} & \mbox{0.721\,(\textcolor{red}{$\uparrow$}0.070)} & \mbox{0.659\,(\textcolor{red}{$\uparrow$}0.066)} & \mbox{0.549\,(\textcolor{red}{$\uparrow$}0.278)} & \mbox{0.464\,(\textcolor{red}{$\uparrow$}0.244)} & \mbox{0.883\,(\textcolor{blue}{$\downarrow$}0.010)} & \mbox{0.791\,(\textcolor{blue}{$\downarrow$}0.016)} & \mbox{0.946\,(\textcolor{red}{$\uparrow$}0.028)} & \mbox{0.898\,(\textcolor{red}{$\uparrow$}0.038)} \\
FixMatch &  & \mbox{0.803\,(\textcolor{red}{$\uparrow$}0.019)} & \mbox{0.714\,(\textcolor{red}{$\uparrow$}0.024)} & \mbox{0.727\,(\textcolor{red}{$\uparrow$}0.059)} & \mbox{0.666\,(\textcolor{red}{$\uparrow$}0.053)} & \mbox{0.538\,(\textcolor{red}{$\uparrow$}0.307)} & \mbox{0.455\,(\textcolor{red}{$\uparrow$}0.270)} & \mbox{0.891\,(\textcolor{red}{$\uparrow$}0.001)} & \mbox{0.804\,(\textcolor{red}{$\uparrow$}0.001)} & \mbox{0.950\,(\textcolor{red}{$\uparrow$}0.002)} & \mbox{0.906\,(\textcolor{red}{$\uparrow$}0.004)} \\
UniMatch~v1 &  & \mbox{0.803\,(\textcolor{red}{$\uparrow$}0.010)} & \mbox{0.715\,(\textcolor{red}{$\uparrow$}0.006)} & \mbox{0.700\,(\textcolor{red}{$\uparrow$}0.014)} & \mbox{0.642\,(\textcolor{red}{$\uparrow$}0.013)} & \mbox{0.539\,(\textcolor{red}{$\uparrow$}0.309)} & \mbox{0.455\,(\textcolor{red}{$\uparrow$}0.263)} & \mbox{0.892\,(\textcolor{blue}{$\downarrow$}0.001)} & \mbox{0.806\,(\textcolor{blue}{$\downarrow$}0.001)} & \mbox{0.950\,(\textcolor{blue}{$\downarrow$}0.006)} & \mbox{0.907\,(\textcolor{blue}{$\downarrow$}0.010)} \\
UniMatch~v2 &  & \mbox{0.822\,(\textcolor{red}{$\uparrow$}0.036)} & \mbox{0.736\,(\textcolor{red}{$\uparrow$}0.033)} & \mbox{0.745\,(\textcolor{red}{$\uparrow$}0.061)} & \mbox{0.685\,(\textcolor{red}{$\uparrow$}0.053)} & \mbox{0.525\,(\textcolor{red}{$\uparrow$}0.309)} & \mbox{0.447\,(\textcolor{red}{$\uparrow$}0.272)} & \mbox{0.888\,(\textcolor{blue}{$\downarrow$}0.004)} & \mbox{0.800\,(\textcolor{blue}{$\downarrow$}0.007)} & \mbox{0.956\,(\textcolor{red}{$\uparrow$}0.006)} & \mbox{0.916\,(\textcolor{red}{$\uparrow$}0.009)} \\
\midrule
VESSA distill & \multirow{5}{*}{\centering 5\%} & 0.818 & 0.730 & 0.793 & 0.728 & 0.570 & 0.480 & 0.835 & 0.718 & 0.939 & 0.886 \\
MT &  & \mbox{0.848\,(\textcolor{red}{$\uparrow$}0.003)} & \mbox{0.767\,(\textcolor{blue}{$\downarrow$}0.001)} & \mbox{0.823\,(\textcolor{red}{$\uparrow$}0.030)} & \mbox{0.767\,(\textcolor{red}{$\uparrow$}0.027)} & \mbox{0.643\,(\textcolor{red}{$\uparrow$}0.063)} & \mbox{0.557\,(\textcolor{red}{$\uparrow$}0.055)} & \mbox{0.892\,(\textcolor{blue}{$\downarrow$}0.012)} & \mbox{0.806\,(\textcolor{blue}{$\downarrow$}0.020)} & \mbox{0.962\,(\textcolor{red}{$\uparrow$}0.002)} & \mbox{0.927\,(\textcolor{red}{$\uparrow$}0.004)} \\
FixMatch &  & \mbox{0.849\,(\textcolor{red}{$\uparrow$}0.012)} & \mbox{0.769\,(\textcolor{red}{$\uparrow$}0.011)} & \mbox{0.826\,(\textcolor{red}{$\uparrow$}0.024)} & \mbox{0.772\,(\textcolor{red}{$\uparrow$}0.022)} & \mbox{0.656\,(\textcolor{red}{$\uparrow$}0.026)} & \mbox{0.570\,(\textcolor{red}{$\uparrow$}0.024)} & \mbox{0.903\,(\textcolor{red}{$\sim$}\textcolor{black}{0.000})} & \mbox{0.824\,(\textcolor{red}{$\uparrow$}0.001)} & \mbox{0.960\,(\textcolor{blue}{$\downarrow$}0.002)} & \mbox{0.924\,(\textcolor{blue}{$\downarrow$}0.003)} \\
UniMatch~v1 &  & \mbox{0.853\,(\textcolor{red}{$\uparrow$}0.013)} & \mbox{0.772\,(\textcolor{red}{$\uparrow$}0.010)} & \mbox{0.836\,(\textcolor{red}{$\uparrow$}0.008)} & \mbox{0.783\,(\textcolor{red}{$\uparrow$}0.008)} & \mbox{0.639\,(\textcolor{red}{$\uparrow$}0.034)} & \mbox{0.555\,(\textcolor{red}{$\uparrow$}0.032)} & \mbox{0.905\,(\textcolor{red}{$\uparrow$}0.001)} & \mbox{0.827\,(\textcolor{red}{$\uparrow$}0.001)} & \mbox{0.961\,(\textcolor{red}{$\sim$}\textcolor{black}{0.000})} & \mbox{0.926\,(\textcolor{red}{$\sim$}\textcolor{black}{0.000})} \\
UniMatch~v2 &  & \mbox{0.861\,(\textcolor{red}{$\uparrow$}0.018)} & \mbox{0.783\,(\textcolor{red}{$\uparrow$}0.018)} & \mbox{0.843\,(\textcolor{red}{$\uparrow$}0.012)} & \mbox{0.789\,(\textcolor{red}{$\uparrow$}0.013)} & \mbox{0.639\,(\textcolor{red}{$\uparrow$}0.013)} & \mbox{0.546\,(\textcolor{red}{$\uparrow$}0.004)} & \mbox{0.912\,(\textcolor{red}{$\uparrow$}0.005)} & \mbox{0.838\,(\textcolor{red}{$\uparrow$}0.008)} & \mbox{0.961\,(\textcolor{red}{$\uparrow$}0.003)} & \mbox{0.925\,(\textcolor{red}{$\uparrow$}0.006)} \\
\midrule
VESSA distill & \multirow{5}{*}{\centering 10\%} & 0.822 & 0.737 & 0.807 & 0.741 & 0.629 & 0.544 & 0.837 & 0.722 & 0.955 & 0.914 \\
MT &  & \mbox{0.855\,(\textcolor{red}{$\uparrow$}0.002)} & \mbox{0.778\,(\textcolor{red}{$\uparrow$}0.003)} & \mbox{0.846\,(\textcolor{red}{$\uparrow$}0.005)} & \mbox{0.791\,(\textcolor{red}{$\uparrow$}0.005)} & \mbox{0.697\,(\textcolor{red}{$\uparrow$}0.018)} & \mbox{0.612\,(\textcolor{red}{$\uparrow$}0.017)} & \mbox{0.907\,(\textcolor{red}{$\sim$}\textcolor{black}{0.000})} & \mbox{0.830\,(\textcolor{red}{$\uparrow$}0.001)} & \mbox{0.966\,(\textcolor{red}{$\uparrow$}0.002)} & \mbox{0.934\,(\textcolor{red}{$\uparrow$}0.003)} \\
FixMatch &  & \mbox{0.858\,(\textcolor{red}{$\uparrow$}0.012)} & \mbox{0.782\,(\textcolor{red}{$\uparrow$}0.013)} & \mbox{0.851\,(\textcolor{red}{$\uparrow$}0.010)} & \mbox{0.797\,(\textcolor{red}{$\uparrow$}0.008)} & \mbox{0.690\,(\textcolor{red}{$\uparrow$}0.018)} & \mbox{0.606\,(\textcolor{red}{$\uparrow$}0.019)} & \mbox{0.903\,(\textcolor{blue}{$\downarrow$}0.007)} & \mbox{0.823\,(\textcolor{blue}{$\downarrow$}0.011)} & \mbox{0.965\,(\textcolor{red}{$\uparrow$}0.002)} & \mbox{0.933\,(\textcolor{red}{$\uparrow$}0.003)} \\
UniMatch~v1 &  & \mbox{0.856\,(\textcolor{blue}{$\downarrow$}0.007)} & \mbox{0.778\,(\textcolor{blue}{$\downarrow$}0.007)} & \mbox{0.857\,(\textcolor{red}{$\uparrow$}0.002)} & \mbox{0.804\,(\textcolor{red}{$\uparrow$}0.003)} & \mbox{0.682\,(\textcolor{red}{$\uparrow$}0.014)} & \mbox{0.600\,(\textcolor{red}{$\uparrow$}0.015)} & \mbox{0.911\,(\textcolor{red}{$\uparrow$}0.004)} & \mbox{0.838\,(\textcolor{red}{$\uparrow$}0.006)} & \mbox{0.966\,(\textcolor{red}{$\sim$}\textcolor{black}{0.000})} & \mbox{0.934\,(\textcolor{red}{$\sim$}\textcolor{black}{0.000})} \\
UniMatch~v2 &  & \mbox{0.869\,(\textcolor{red}{$\uparrow$}0.012)} & \mbox{0.792\,(\textcolor{red}{$\uparrow$}0.012)} & \mbox{0.857\,(\textcolor{red}{$\uparrow$}0.001)} & \mbox{0.804\,(\textcolor{red}{$\uparrow$}0.002)} & \mbox{0.700\,(\textcolor{red}{$\uparrow$}0.027)} & \mbox{0.613\,(\textcolor{red}{$\uparrow$}0.024)} & \mbox{0.907\,(\textcolor{blue}{$\downarrow$}0.003)} & \mbox{0.830\,(\textcolor{blue}{$\downarrow$}0.005)} & \mbox{0.965\,(\textcolor{red}{$\uparrow$}0.004)} & \mbox{0.932\,(\textcolor{red}{$\uparrow$}0.006)} \\
\bottomrule
\end{tabular}%
}\vspace{-8pt}
\end{table*}

Table~\ref{tab:vessa_frameworks} evaluates the framework-level robustness of VESSA by integrating it with Mean Teacher, FixMatch, and UniMatch~v1/v2; we also report ``VESSA distill,'' which directly distills VESSA pseudo-labels without an additional SSL framework. VESSA improves Dice and mIoU in nearly all framework--dataset--label combinations, with the largest gains at 1\% labels and on the hardest targets. For example, at 1\% labels, it raises Mean Teacher Dice from 0.271 to 0.549 on BUSI and from 0.651 to 0.721 on AbdomenCT. These consistent gains across distinct pseudo-labeling and consistency designs support VESSA's plug-and-play applicability.

Overall, template-guided VESSA supervision is most valuable under extreme label scarcity and on datasets with small structures, weak boundaries, or strong domain variability, where self-training provides limited external guidance. As more labels become available, the task-specific student strengthens and VESSA's contribution becomes complementary rather than dominant.

\textbf{Qualitative results.}
Fig.~\ref{fig:qual} presents a multi-class qualitative comparison at 1\% labels across the five Stage-2 datasets. VESSA-assisted SSL produces more coherent masks with smoother, anatomically faithful boundaries, whereas competing baselines often exhibit fragmented regions or boundary leakage. The advantage is most apparent on ACDC, AbdomenCT, and BUSI, where VESSA recovers more complete structures and precise boundaries for small or low-contrast targets; qualitative differences remain visible even on the quantitatively near-saturated STS-2D Tooth and Shenzhen CXR datasets.

\subsection{Ablation Study}
\label{sec:ablation}

All ablations use UniMatch~v2 on ACDC under the 1\%, 5\%, and 10\% labeled settings.

\begin{table*}[!t]
\centering
\begin{minipage}[t]{0.490\textwidth}
\centering
\captionof{table}{Ablation of the Stage-2 training objective. Every row except ``\((\mathcal{L}_{\mathrm{sup}} + \mathcal{L}_{\mathrm{u}}\)) only'' adds a VESSA guidance term to the baseline semi-supervised objective.}
\label{tab:ablation_loss}
\resizebox{\linewidth}{!}{
\begin{tabular}{l|cc|cc|cc}
\toprule
\multirow{2}{*}{Objective}
& \multicolumn{2}{c|}{1\%}
& \multicolumn{2}{c|}{5\%}
& \multicolumn{2}{c}{10\%} \\
\cline{2-7}
& Dice & mIoU & Dice & mIoU & Dice & mIoU \\
\midrule
$(\mathcal{L}_{\mathrm{sup}} + \mathcal{L}_{\mathrm{u}})$ only
& 0.786 & 0.703
& 0.843 & 0.765
& 0.857 & 0.780 \\

$\mathcal{L}_{\mathrm{CE}}+(\mathcal{L}_{\mathrm{sup}} + \mathcal{L}_{\mathrm{u})}$
& \underline{0.809} & \underline{0.722}
& \underline{0.845} & \underline{0.764}
& 0.856 & 0.777 \\

$\mathcal{L}_{\mathrm{CE}}+\mathcal{L}_{\mathrm{KL}}+(\mathcal{L}_{\mathrm{sup}} + \mathcal{L}_{\mathrm{u}})$
& \textbf{0.822} & \textbf{0.736}
& \textbf{0.861} & \textbf{0.783}
& \textbf{0.869} & \textbf{0.792} \\

$\mathcal{L}_{\mathrm{KL}}+(\mathcal{L}_{\mathrm{sup}} + \mathcal{L}_{\mathrm{u}})$
& 0.807 & 0.719
& 0.840 & 0.757
& \underline{0.859} & \underline{0.780} \\
\bottomrule
\end{tabular}}
\end{minipage}\hfill
\begin{minipage}[t]{0.487\textwidth}
\centering
\captionof{table}{Ablation of the reference and memory mechanisms. ``w/o reference'' removes the template, ``w/o memory'' removes the memory module, and ``w/ ref \& mem'' denotes VESSA.}
\label{tab:ablation_refmem}
{\renewcommand{\arraystretch}{1.00}
\resizebox{0.845\linewidth}{!}{
\begin{tabular}{l|cc|cc|cc}
\toprule
\multirow{2}{*}{Labeled} & \multicolumn{2}{c|}{w/o reference} & \multicolumn{2}{c|}{w/o memory} & \multicolumn{2}{c}{w/ ref \& mem} \\
\cline{2-7}
\rule[-0.3ex]{0pt}{2.5ex} & Dice & mIoU & Dice & mIoU & Dice & mIoU \\
\midrule
1\%  & 0.786 & 0.700 & 0.784 & 0.692 & \textbf{0.822} & \textbf{0.736} \\
5\%  & 0.821 & 0.742 & 0.834 & 0.751 & \textbf{0.861} & \textbf{0.783} \\
10\% & 0.832 & 0.754 & 0.849 & 0.768 & \textbf{0.869} & \textbf{0.792} \\
\bottomrule
\end{tabular}}}
\end{minipage}
\vspace{-1.2\baselineskip}
\end{table*}

\paragraph{Effectiveness of the reference mechanism}
The reference mechanism lets VESSA exploit a small bank of gold-standard templates. Table~\ref{tab:vessa_stagewise_performance} compares matched-template VESSA (5/10/20\%) with the reference-free variant (``No-ref'') at the foundation-model level. Templates improve Dice on every dataset and by more than 0.3 on several (\eg, LGG and BraTS), demonstrating their importance for accurate few-shot segmentation across modalities and anatomies.

The reference is equally important for Stage-2 SSL, where scarce labeled samples form the template bank. In Table~\ref{tab:ablation_refmem}, removing the reference forces VESSA to generate pseudo-labels using only the target segmentation classes in the text prompt, reducing their quality and downstream performance. Full VESSA with reference and memory yields consistently higher Dice and mIoU, improving ACDC Dice by 3.6--4.0 points across label budgets. Thus, template evidence supports both few-shot inference and label-aware SSL guidance.

\paragraph{Effectiveness of the memory mechanism}
Beyond selecting a matched template, VESSA encodes the template image and its annotation into a template-embedded memory that the target features attend to during decoding. To assess this component, the ``w/o memory'' column of Table~\ref{tab:ablation_refmem} evaluates VESSA with the memory module disabled while keeping the reference/template overlay to the vision--language backbone unchanged. 
Removing the memory path eliminates direct template-memory conditioning while retaining the reference overlay used by the vision–language backbone; the resulting drop in downstream Dice and mIoU shows that the template-embedded memory provides complementary spatial grounding that the reference alone does not, and is an essential part of VESSA's guidance.

\paragraph{Effectiveness of the Stage-2 training objective}
Finally, we study the objective used to distill VESSA guidance during Stage-2 training. The overall Stage-2 loss includes labeled supervision \(\mathcal{L}_{\mathrm{sup}}\) and the weak-to-strong consistency loss \(\mathcal{L}_{\mathrm{u}}\). It also includes a VESSA guidance term with two components. The hard cross-entropy loss \(\mathcal{L}_{\mathrm{CE}}\) uses the VESSA pseudo-label, while the soft Kullback--Leibler loss \(\mathcal{L}_{\mathrm{KL}}\) uses the full VESSA probability map (Eq.~\ref{eq:lvessa}). As reported in Table~\ref{tab:ablation_loss}, adding the VESSA guidance clearly outperforms the baseline semi-supervised training, and combining the two guidance terms gives the best results across all labeled ratios: \(\mathcal{L}_{\mathrm{CE}}\) anchors the student to VESSA's confident predictions, while \(\mathcal{L}_{\mathrm{KL}}\) transfers its full class-probability structure. Therefore, the two components are complementary. We adopt an initial weighting, \(0.5\,\mathcal{L}_{\mathrm{CE}}+0.5\,\mathcal{L}_{\mathrm{KL}}+\mathcal{L}_{\mathrm{sup}} + \mathcal{L}_{\mathrm{u}}\) (ratio 1:1:2:2), as our default, since we empirically observe that varying the CE-to-KL ratio between roughly 0.25 and 0.75 makes little difference to the final accuracy.
\FloatBarrier
\section{Conclusion}
\label{sec:conclusion}

In this work, we introduced VESSA, a vision--language enhanced medical segmentation foundation model designed to bridge flexible text-prompted few-shot segmentation and semi-supervised medical image segmentation. Unlike conventional SSL methods that mainly rely on perturbation consistency and self-generated pseudo-labels, VESSA incorporates scarce gold-standard templates through reference matching and template-embedded memory, providing external anatomical and semantic guidance under limited annotation. Extensive experiments on multiple benchmark medical datasets under label-scarce settings confirm the robustness and effectiveness of our approach.
By combining text-based semantic flexibility, few-shot template guidance, and seamless integration into semi-supervised learners, VESSA improves label-efficient medical image segmentation across heterogeneous modalities and anatomical targets.
In future work, we aim to scale up VESSA's vision--language backbone and
pre-training data scope, to leverage reinforcement learning to further improve
the generalization of the pre-trained Stage-1 model, and to explore more
effective techniques for strengthening Stage-2 semi-supervised segmentation.

\bibliographystyle{IEEEtran}
\bibliography{main}

\end{document}